%% file: main.tex
\documentclass[10pt,twocolumn,letterpaper]{article}

\usepackage{iccv}
\usepackage{times}
\usepackage{epsfig}
\usepackage{epstopdf}
\usepackage{enumitem}
\usepackage{graphicx}
\usepackage{amsmath}
\usepackage{amssymb}
\usepackage{booktabs}
\usepackage{array}
\usepackage{xcolor}
\usepackage{multirow}
\usepackage{caption}
\usepackage{subcaption}
\usepackage{float}

\usepackage[pagebackref=true,breaklinks=true,letterpaper=true,colorlinks,bookmarks=false]{hyperref}

\input{math_definition}
\newcommand{\trainS}{base classes\xspace}
\newcommand{\valS}{validation classes\xspace}
\newcommand{\testS}{novel classes\xspace}
\newcommand{\CNN}{Conv-4\xspace}
\newcommand{\ResNet}{ResNet\xspace}

\newcommand*{\rowstyle}[1]{
	\gdef\@rowstyle{#1}%
	\@rowstyle\ignorespaces%
}
\newcolumntype{=}{
	>{\gdef\@rowstyle{}}%
}
\newcolumntype{+}{
	>{\@rowstyle}%
}
\newcommand{\basePlain}{SimpleShot (UN)}
\newcommand{\baseFair}{SimpleShot (L2N)}
\newcommand{\baseBest}{SimpleShot (CL2N)}

\makeatletter
\def\and{%
  \end{tabular}%
  \hskip 0.935em \@plus.17fil\relax
  \begin{tabular}[t]{c}}
\makeatother

\iccvfinalcopy 

\ificcvfinal\fi
\begin{document}

\title{SimpleShot: Revisiting Nearest-Neighbor Classification for Few-Shot Learning}

\author{Yan Wang\\
Cornell University\\
{\tt\small yw763@cornell.edu}
\and
Wei-Lun Chao\\
Ohio State University\\
{\tt\small chao.209@osu.edu}
\and
Kilian Q. Weinberger\\
Cornell University\\
{\tt\small kqw4@cornell.edu}
\and
Laurens van der Maaten\\
Facebook AI Research\\
{\tt\small lvdmaaten@fb.com}
}

\maketitle

\begin{abstract}
Few-shot learners aim to recognize new object classes based on a small number of labeled training examples. To prevent overfitting, state-of-the-art few-shot learners use meta-learning on convolutional-network features and perform classification using a nearest-neighbor classifier. This paper studies the accuracy of nearest-neighbor baselines without meta-learning. Surprisingly, we find simple feature transformations suffice to obtain competitive few-shot learning accuracies. For example, we find that a nearest-neighbor classifier used in combination with mean-subtraction and L2-normalization outperforms prior results in three out of five settings on the miniImageNet dataset.

\end{abstract}

\input{introduction}
\input{approach}
\input{experiments}
\input{discussion}

\vspace{-1ex}
\paragraph{Acknowledgments}
{\small
The authors thank Han-Jia Ye for helpful discussions. Y.W. and K.Q.W. are supported by grants from the NSF (III-1618134, III-1526012, IIS-1149882, IIS-1724282, and TRIPODS-1740822), the Bill and Melinda Gates Foundation, and the Cornell Center for Materials Research with funding from the NSF MRSEC program (DMR-1719875); and are also supported by Zillow, SAP America Inc., and Facebook.

\clearpage

\bibliographystyle{ieee}
\bibliography{egbib}
}
\clearpage

\appendix
\input{suppl_content}
\end{document}

%% file: math_definition.tex
\usepackage{amsmath,amsfonts}
\usepackage{amssymb,amsopn}
\usepackage{bm} 

\usepackage{amssymb}
\usepackage{amsmath,amsfonts}
\usepackage{amsthm,amsopn}

\usepackage{bm} 

\newcommand{\vct}[1]{\boldsymbol{#1}} 

\newcommand{\field}[1]{\mathbb{#1}}
\newcommand{\R}{\field{R}} 

\newcommand{\ProbOpr}[1]{\mathbb{#1}}

\newcommand{\expect}[2]{%
\ifthenelse{\equal{#2}{}}{\ProbOpr{E}_{#1}}
{\ifthenelse{\equal{#1}{}}{\ProbOpr{E}\left[#2\right]}{\ProbOpr{E}_{#1}\left[#2\right]}}} 
\newcommand{\var}[2]{%
\ifthenelse{\equal{#2}{}}{\ProbOpr{VAR}_{#1}}
{\ifthenelse{\equal{#1}{}}{\ProbOpr{VAR}\left[#2\right]}{\ProbOpr{VAR}_{#1}\left[#2\right]}}} 

\DeclareMathOperator{\argmin}{arg\,min}

\newcommand{\vx}{{\vct{x}}}

\newcommand{\eat}[1]{}

%% file: introduction.tex

\section{Introduction}
\label{sec:introduction}
The human visual system has an ability to recognize new visual classes (for instance, greebles~\cite{gauthier1998}) based on a few examples that is, currently, unmatched by computer vision. The development of computer-vision systems that can perform such \emph{few-shot learning}~\cite{fei2006one,snell2017prototypical,vinyals2016matching} is important, \emph{e.g.}, for developing systems that can recognize the millions of natural or man-made classes that appear in the world~\cite{vanhorn2017}.

Few-shot learning is generally studied in a learning setting in which the visual-recognition system is first trained to recognize a collection of \emph{\trainS} from a large number of training examples. Subsequently, the system receives a small number of training examples (so-called \emph{``shots''}) for a few novel visual classes that it needs to recognize thereafter. In order to be robust to overfitting, a successful few-shot learning model must efficiently re-use what it learned from training on the \trainS{} for the novel classes.

\begin{figure}
\vspace{5ex}
	\centering
	~ 
	\begin{subfigure}[b]{\linewidth}
		\includegraphics[width=\linewidth,trim={0 0 0 0.68cm},clip]{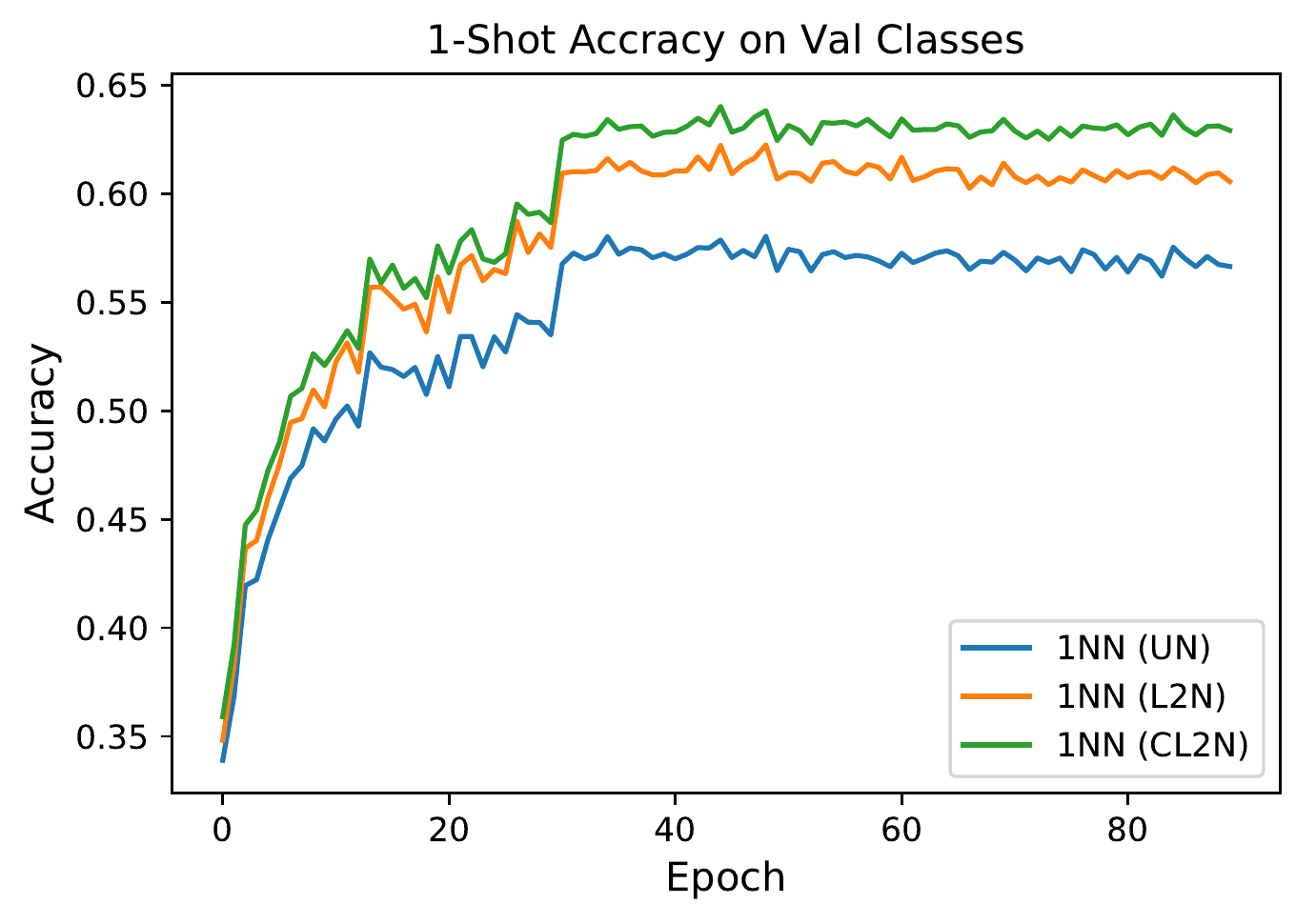}
	\end{subfigure}
	\caption{\small \textbf{Feature transformations matter in few-shot learning using nearest neighbors.} We train a DenseNet on \emph{mini}ImageNet and use the learned features to perform few-shot learning using a nearest-neighbor classifier with Euclidean distance. We measure the one-shot five-way accuracy on 10,000 tasks sampled from the validation classes during training. We compare un-normalized (UN), L2-normalized (L2N), and centered L2-normalized (CL2N) features. CL2N features outperform UN features, highlighting the importance of feature transformations in few-shot learning.}
	\label{fig:concept}
\end{figure}

Many current few-shot learners extract image features using a convolutional network, and use a combination of meta-learning and nearest-neighbor classification to perform the recognition~\cite{oreshkin2018tadam,vinyals2016matching,snell2017prototypical,sung2018learning,ye2018learning}. Prior studies suggest that using meta-learning outperforms ``vanilla'' nearest neighbor classification~\cite{ravi2017optimization,snell2017prototypical}.

This study challenges the status quo by demonstrating that nearest-neighbor classifiers can achieve state-of-the-art performance on popular few-shot learning benchmarks without meta-learning. Specifically, we find that applying simple feature transformations on the features before nearest-neighbor classification leads to very competitive few-shot learning results. For example, we find that a nearest-neighbor classifier that uses DenseNet features~\cite{huang2017densely} to which mean subtraction and L2-normalization are applied outperforms a long list~\cite{bauer2017discriminative,finn2017model,finn2018probabilistic,garcia2018few,gidaris2018dynamic,gordon2019meta,grant2018recasting,jiang2019learning,lee2018meta,mishra2018simple,munkhdalai2018rapid,nichol2018first,oreshkin2018tadam,ravi2017optimization,rusu2019meta,qiao2018few,snell2017prototypical,sung2018learning,triantafillou2017few,vinyals2016matching,yoon2018bayesian} of recent, arguably more complex few-shot learning approaches on the popular \emph{mini}ImageNet~\cite{vinyals2016matching} and \emph{tiered}ImageNet~\cite{ren2018meta} benchmarks (see Table~\ref{tbMain_1} and~\ref{tbMain_2}). These observations generalize to other convolutional network architectures~\cite{he2016deep,howard2017mobilenets,zagoruyko2016wide}. We refer to our few-shot learner as SimpleShot. We hope to re-establish nearest-neighbor classification as an obvious but competitive baseline for few-shot learning.

%% file: approach.tex
\section{Nearest Neighbors for Few-Shot Learning}
\label{sec:apprach}

Denoting an image by $\mathbf{I}$, we assume we are given a training set, $\mathcal{D}_\text{base} = \{(\mathbf{I}_1, y_1),  \dots, (\mathbf{I}_N, y_N)\}$, that contains $N$ labeled images from $A$ \trainS; that is, $y_n\in\{1,\dots, A\}$. Furthermore, we assume we are given a support set $\mathcal{D}_\text{support}$ of labeled images from $C$ \testS, where each novel class has $K$ examples. The goal of few-shot learning is to construct a model that accurately recognizes the $C$ novel classes. This learning setting is referred to as the $K$-shot $C$-way setting. 

We study a few-shot learner based on nearest-neighbor classification, called SimpleShot. The nearest-neighbor classifier operates on features $\vx\!\in\!\R^D$ that were extracted from image $\textbf{I}$ using a convolutional network $f_\theta(\mathbf{I})$ with  parameters $\theta$. The feature-producing convolutional network, $f_\theta(\mathbf{I})$, is trained to minimize the loss of a linear classifier (with $\mathbf{W}\in\R^{D\times A}$ in the last network layer) on $\mathcal{D}_\text{base}$:
\begin{equation*}
\argmin_{\theta, \mathbf{W}} \sum_{(\mathbf{I}, y) \in \mathcal{D}_\text{base}} \ell(\mathbf{W}^\top f_\theta(\mathbf{I}), y),
\end{equation*}
where the loss function $\ell$ is selected to be the cross-entropy loss. The convolutional network and the linear classifier are trained jointly using stochastic gradient descent.

\vspace{-1ex}\paragraph{Nearest Neighbor Rule.} Once the feature extraction network, $f_\theta$, is trained on the \trainS{}, we access images exclusively in feature space and consider all subsequent images as readily provided in feature space. 
For simplicity of notation, we denote $\vx=f_\theta(\mathbf{I})$ as an image in feature space. 
In this space we perform nearest-neighbor classification using some distance measure, $d(\vx, \vx') \in \R^+_0$. 
We first consider the \textbf{one-shot} setting, that is, the setting in which $\mathcal{D}_\text{support}$ contains only $K=1$ labeled example for each of the $C$ classes:  $\mathcal{D}_\text{support} = \{(\hat{\vx
}_1, 1), \dots, (\hat{\vx}_C, C)\}$, where we use the notation $\hat{\vx}$ to distinguish images in the novel $C$ classes from images $\vx$ in $\mathcal{D}_\text{base}$. The nearest-neighbor rule assigns the label of the most similar support image (in feature space) to a test image $\hat{\vx}$:
\begin{equation}
y(\hat{\vx}) = \argmin_{c\in\{1,\cdots,C\}} d(\hat{\vx}, \hat{\vx}_c). \label{e_NN}
\end{equation}
In \textbf{multi-shot} settings, we use a nearest-centroid approach. Specifically, we compute the averaged feature vector (centroid) for each class in $\mathcal{D}_\text{support}$ and treat each of the centroids as a one-shot example for the corresponding class. We then apply Equation~\ref{e_NN} on the centroids.

\subsection{Feature Transformations}
In this study, we use the Euclidean distance, $d(\hat{\vx}, \hat{\vx}') = \|\hat{\vx} - \hat{\vx}'\|_2$, as the distance measure for nearest-neighbors classification.
We only consider two feature transformations that are well-established and may be considered trivial but, empirically, we find that they can have a positive effect on the accuracy of the SimpleShot few-shot learner. 

\paragraph{Centering.}
We compute the mean feature vector on the \trainS{}, 
$\bar{\vx} = \frac{1}{|\mathcal{D}_\text{base}|} \sum_{\vx\in \mathcal{D}_\text{base}} \vx$, 
and subtract it from a feature vector $\hat{\vx}$ to normalize it: $\hat{\vx}\leftarrow\hat{\vx} - \bar{\vx}$. Centering (or mean subtraction) in itself does not alter Euclidean distances between feature vectors, but can become effective in combination with L2-normalization. 
\vspace{-2mm}
\paragraph{L2-normalization (L2N).}~~~Given a feature vector $\hat{\vx}$, we normalize it to have unit $\ell_2$ norm: $\hat{\vx}\leftarrow\frac{\hat{\vx}}{\|\hat{\vx}\|_2}$.

%% file: experiments.tex
\vspace{-2mm}
\section{Experiments}
\label{sec:experiments}

Following prior work, we measure the efficacy of feature transformations in nearest-neighbor classifiers for few-shot learning in a series of image-recognition experiments.\footnote{Code at \url{https://github.com/mileyan/simple_shot}.}

\subsection{Experimental Setup}

\paragraph{Datasets.} We experiment on three image datasets.

The \textbf{\emph{mini}ImageNet dataset}~\cite{vinyals2016matching} is a subset of ImageNet~\cite{RussakovskyDSKS15ImageNet} that is commonly used to study few-shot learning. The dataset contains 100 classes and has a total of 600 examples per class. Following~\cite{ravi2017optimization} and subsequent work, we split the dataset to have 64 \trainS, 16 \valS, and 20 \testS. Following~\cite{vinyals2016matching} and subsequent studies, we resize the images to $84 \times 84$ pixels via rescaling and center cropping.

We also perform experiments on the \textbf{\emph{tiered}ImageNet dataset}~\cite{ren2018meta}, which is also constructed from ImageNet but contains 608 classes. The dataset is split into 351, 97, and 160 classes for base, validation, and \testS, respectively. The class split is performed using WordNet~\cite{Miller95} to ensure
that all the \trainS are semantically unrelated to the \testS. Again, we resize images to $84 \times 84$ pixels.

Following~\cite{oreshkin2018tadam}, we also perform experiments on the \textbf{CIFAR-100}~\cite{krizhevsky2009learning} dataset, which contains 100 image classes. Each of the classes in the dataset has 600 images of size $32 \times 32$ pixels. We follow~\cite{oreshkin2018tadam} and split the classes into 
60 base, 20 validation, and 20 novel classes.



\begin{table}[t]
	\centering
	\small {
		\caption{\small Average accuracy (in \%; measured over 600/10,000 rounds$^\star$) of one-shot and five-shot classifiers for five-way classification on \emph{mini}ImageNet; higher is better. The best result of each network architecture of each column is in \textbf{bold} font. Results of our approaches are in {\color{blue} blue}.  Best viewed in color.} \label{tbMain_1}
		\resizebox{\linewidth}{!}{  
			\begin{tabular}{=l +l +c +c}
				\toprule
				\textbf{Approach} & \textbf{Network} & \textbf{One shot} & \textbf{Five shots} \\ \midrule
				Meta LSTM~\cite{ravi2017optimization}& \CNN & 43.44 $\pm$ 0.77 & 60.60 $\pm$ 0.71 \\
				MatchingNet~\cite{vinyals2016matching} & \CNN & 43.56 $\pm$ 0.84 & 55.31 $\pm$ 0.73 \\
				MAML~\cite{finn2017model} & \CNN & 48.70 $\pm$ 1.84 & 63.11 $\pm$ 0.92 \\
				LLAMA~\cite{grant2018recasting} & \CNN & 49.40 $\pm$ 1.83 & -- \\
				ProtoNet~\cite{snell2017prototypical} & \CNN & 49.42 $\pm$ 0.78 & 68.20 $\pm$ 0.66 \\
				Reptile~\cite{nichol2018first}& \CNN & 49.97 $\pm$ 0.32 & 65.99 $\pm$ 0.58\\
				PLATIPUS~\cite{finn2018probabilistic}& \CNN & 50.13 $\pm$ 1.86& -- \\
				mAP-SSVM~\cite{triantafillou2017few} & \CNN & 50.32 $\pm$ 0.80 & 63.94 $\pm$ 0.72\\
				GNN~\cite{garcia2018few} & \CNN & 50.33 $\pm$ 0.36& 66.41 $\pm$ 0.63\\
				RelationNet~\cite{sung2018learning} & \CNN & 50.44 $\pm$ 0.82 & 65.32 $\pm$ 0.70 \\
				Meta SGD~\cite{li2017meta} & \CNN &  50.47 $\pm$ 1.87 & 64.03 $\pm$ 0.94 \\ 
				MTNet~\cite{lee2018meta}& \CNN & 51.70 $\pm$ 1.84 & -- \\
				Qiao \emph{et al.}~\cite{qiao2018few} &\CNN  & 54.53 $\pm$ 0.40 & 67.87 $\pm$ 0.20\\
				FEAT~\cite{ye2018learning} & \CNN & \textbf{55.15 $\pm$ 0.20} & \textbf{71.61 $\pm$ 0.16} \\
				\rowstyle{\color{blue}}
				\basePlain & \CNN & 33.17 $\pm$ 0.17 & 63.25 $\pm$ 0.17\\
				\rowstyle{\color{blue}}
				\baseFair & \CNN & 48.08 $\pm$ 0.18 & 66.49 $\pm$ 0.17 \\
				\rowstyle{\color{blue}}
				\baseBest & \CNN & 49.69 $\pm$ 0.19  & 66.92 $\pm$ 0.17 \\
				\midrule
				MAML~\cite{finn2017model}$^\dagger$ & \ResNet-18 & 49.61 $\pm$ 0.92 & 65.72 $\pm$ 0.77\\
				Chen~\emph{et al.}~\cite{chen2019closer} & \ResNet-18 &  51.87 $\pm$ 0.77 & 75.68 $\pm$ 0.63 \\
				RelationNet~\cite{sung2018learning}$^\dagger$ & \ResNet-18 & 52.48 $\pm$ 0.86 &  69.83 $\pm$ 0.68 \\
				MatchingNet~\cite{vinyals2016matching}$^\dagger$ & \ResNet-18 & 52.91 $\pm$ 0.88 & 68.88 $\pm$ 0.69 \\
				ProtoNet~\cite{snell2017prototypical}$^\dagger$ & \ResNet-18 &  54.16 $\pm$ 0.82 & 73.68 $\pm$ 0.65 \\
				Gidaris \emph{et al.}~\cite{gidaris2018dynamic} & \ResNet-15 & 55.45 $\pm$ 0.89 & 70.13 $\pm$ 0.68 \\
				SNAIL~\cite{mishra2018simple} & \ResNet-15 & 55.71 $\pm$ 0.99 & 68.88 $\pm$ 0.92 \\
				Bauer \emph{et al.}~\cite{bauer2017discriminative}& \ResNet-34 & 56.30 $\pm$ 0.40 & 73.90 $\pm$ 0.30 \\
				adaCNN~\cite{munkhdalai2018rapid} & \ResNet-15 & 56.88 $\pm$ 0.62 & 71.94 $\pm$ 0.57 \\
				TADAM~\cite{oreshkin2018tadam} & \ResNet-15 & 58.50 $\pm$ 0.30& 76.70 $\pm$ 0.30\\
				CAML~\cite{jiang2019learning} & \ResNet-12 & 59.23 $\pm$ 0.99 &72.35 $\pm$ 0.71\\
				\rowstyle{\color{blue}}
				\basePlain & \ResNet-10 & 54.45 $\pm$ 0.21 & 76.98 $\pm$ 0.15\\
				\rowstyle{\color{blue}}
				\baseFair & \ResNet-10 & 57.85 $\pm$ 0.20 & 78.73 $\pm$ 0.15\\
				\rowstyle{\color{blue}}
				\baseBest & \ResNet-10 & 60.85 $\pm$ 0.20 & 78.40 $\pm$ 0.15 \\
				\rowstyle{\color{blue}}
				\basePlain & \ResNet-18 & 56.06 $\pm$ 0.20  & 78.63 $\pm$ 0.15 \\
				\rowstyle{\color{blue}}
				\baseFair & \ResNet-18 & 60.16 $\pm$ 0.20 & 79.94 $\pm$ 0.14\\
				\rowstyle{\color{blue}}
				\baseBest & \ResNet-18 & \textbf{62.85 $\pm$ 0.20} & \textbf{80.02 $\pm$ 0.14}\\
				\midrule
				Qiao \emph{et al.}~\cite{qiao2018few} &WRN  & 59.60 $\pm$ 0.41 & 73.74 $\pm$ 0.19\\
				MatchingNet~\cite{vinyals2016matching}$^\sharp$ &WRN & 64.03 $\pm$ 0.20& 76.32 $\pm$ 0.16\\
				ProtoNet~\cite{snell2017prototypical}$^\sharp$ & WRN & 62.60 $\pm$ 0.20& 79.97 $\pm$ 0.14\\
				LEO~\cite{rusu2019meta} & WRN & 61.76 $\pm$ 0.08 & 77.59 $\pm$ 0.12\\
				FEAT~\cite{ye2018learning} & WRN & \textbf{65.10 $\pm$ 0.20} & \textbf{81.11 $\pm$ 0.14} \\
				\rowstyle{\color{blue}}
				\basePlain & WRN & 57.26 $\pm$ 0.21 & 78.99 $\pm$ 0.14 \\
				\rowstyle{\color{blue}}
				\baseFair & WRN & 61.22 $\pm$ 0.21 & 81.00 $\pm$ 0.14 \\
				\rowstyle{\color{blue}}
				\baseBest & WRN & 63.50 $\pm$ 0.20  & 80.33 $\pm$ 0.14 \\
				\midrule
				\rowstyle{\color{blue}}
				\basePlain & MobileNet & 55.70 $\pm$ 0.20  & 77.46 $\pm$ 0.15 \\
				\rowstyle{\color{blue}}
				\baseFair & MobileNet  & 59.43 $\pm$ 0.20  & 78.00 $\pm$ 0.15\\
				\rowstyle{\color{blue}}
				\baseBest & MobileNet  & \textbf{61.30 $\pm$ 0.20}  & \textbf{78.37 $\pm$ 0.15} \\
				\midrule
				\rowstyle{\color{blue}}
				\basePlain & DenseNet & 57.81 $\pm$ 0.21 & 80.43 $\pm$ 0.15\\
				\rowstyle{\color{blue}}
				\baseFair & DenseNet  & 61.49 $\pm$ 0.20  & 81.48 $\pm$ 0.14\\
				\rowstyle{\color{blue}}
				\baseBest & DenseNet  & \textbf{64.29 $\pm$ 0.20}  & \textbf{81.50 $\pm$ 0.14}\\
				\bottomrule
			\end{tabular}
		}
		\begin{flushleft}
			$^\dagger$: Results reported in~\cite{chen2019closer}. \hspace{5pt} $^\sharp$: Results reported in~\cite{ye2018learning}. \\
			$^\star$: \cite{rusu2019meta, ye2018learning} and our results are averaged over 10,000 rounds.\\
		\end{flushleft}
	}
\end{table}  

\vspace{-1ex}
\paragraph{Evaluation protocol.}
Following~\cite{rusu2019meta}, we measure the accuracy of SimpleShot and the other few-shot learners by drawing 10,000 $K$-shot $C$-way tasks from the \testS: each task has $C$ \testS and $K$ labeled (support) images and 15 test (query) images per class. Following prior work, we focus on one-shot and five-shot, five-way tasks.

We average observed accuracies over all test images and over all the tasks, and report the resulting average accuracy and $95\%$ confidence interval. 

\vspace{-1ex}
\paragraph{Model and implementation details.}
We evaluate our methods using five different convolutional-network architectures as the basis for the feature-generating function $f_\theta(\mathbf{I})$. We study five different network architectures: 
\begin{itemize}
	\itemsep-0.5em
	\item \textbf{Four-layer convolutional networks (Conv-4)}: We follow~\cite{snell2017prototypical,vinyals2016matching} to implement this baseline model.
	\item \textbf{Wide residual networks (WRN-28-10)}~\cite{zagoruyko2016wide}: We follow~\cite{rusu2019meta} and use the architecture with 28 convolutional layers and a widening factor of 10.
	\item \textbf{Dense convolutional networks (DenseNet-121)}~\cite{huang2017densely}: We use the standard 121-layer architecture but remove the first two down-sampling layers (\emph{i.e.}, we set their stride to $1$) and change the first convolutional layer to use a kernel of size $3 \times 3$ (rather than $7 \times 7$) pixels.
	\item \textbf{Residual networks (ResNet-10/18)}~\cite{he2016deep}:
	We use the standard 18-layer architecture but we remove the first two down-sampling layers and we change the first convolutional layer to use a kernel of size $3 \times 3$ (rather than $7 \times 7$) pixels. Our ResNet-10 contains 4 residual blocks; the ResNet-18 contains 8 blocks.
	\item \textbf{MobileNet}~\cite{howard2017mobilenets}: We use the standard architecture for ImageNet~\cite{RussakovskyDSKS15ImageNet} but, again, we remove the first two down-sampling layers from the network.	
\end{itemize}


We train all networks for 90 epochs from scratch using stochastic gradient descent to minimize the cross-entropy loss of $A$-way classification ($A$ is the number of \trainS). We perform the data augmentation proposed in~\cite{he2016deep}. We set the initial learning rate to $0.1$ and use a batch size of $256$ images. On \emph{mini}ImageNet, We shrink the learning rate by $10$ at 45 and 66 epoch respectively. On \emph{tiered}ImageNet, we divide the learning rate by $10$ after every 30 epochs. We perform early stopping according to the one-shot five-way accuracy (measured using \baseFair) on the \valS.

\vspace{-1ex}
\paragraph{Feature transformations.}
\vskip -7.5pt
We evaluate the effectiveness of three feature transformations in our experiments:
\vspace{-0.5ex}
\begin{itemize}
	\itemsep-0.5em
	\item \textbf{UN}: Unnormalized features. 
	\item \textbf{L2N}: L2-normalized features.
	\item \textbf{CL2N}: Centered and then L2-normalized features.
\end{itemize}
\vspace{-1ex}
These transforms are followed by nearest-neighbor classification using the Euclidean distance measure. 

\vspace{-1ex}
\paragraph{Comparison.}
We compare our baselines to a range of state-of-the-art few-shot learners~\cite{bauer2017discriminative,finn2017model,finn2018probabilistic,garcia2018few,gidaris2018dynamic,grant2018recasting,jiang2019learning,lee2018meta,mishra2018simple,munkhdalai2018rapid,nichol2018first,oreshkin2018tadam,ravi2017optimization,rusu2019meta,qiao2018few,snell2017prototypical,sung2018learning,triantafillou2017few,vinyals2016matching,ye2018learning}. We do not compare to approaches that were developed for semi-supervised and transductive learning settings, as such approaches use the statistics of query examples or statistics across the few-shot tasks. We note that the network architectures used in prior studies may have slight variations; we have tried our best to eliminate the effect of such variations on our observations as much as possible.\footnote{For example, we report results for ResNet-10 models because it is the shallowest ResNet architecture used in prior work on few-shot learning.}

\begin{table}[t]
	\centering
	\small {
		\caption{\small Average accuracy (in \%; measured over 600/10,000 rounds$^\star$) of one-shot and five-shot classifiers for five-way classification on \emph{tiered}ImageNet; higher is better. The best result of each network architecture of each column is in \textbf{bold} font. Results of our approach are in {\color{blue} blue}. Best viewed in color.} \label{tbMain_2}
		\resizebox{\linewidth}{!}{  
			\begin{tabular}{=l +l +c +c}
				\toprule
				\textbf{Approach} & \textbf{Network} & \textbf{One shot} & \textbf{Five shots} \\ \midrule
				Reptile~\cite{nichol2018first}$^\sharp$ & \CNN & 48.97 $\pm$ 0.21& 66.47 $\pm$ 0.21\\
				ProtoNet~\cite{snell2017prototypical}$^\sharp$ & \CNN & \textbf{53.31 $\pm$ 0.89} & \textbf{72.69 $\pm$ 0.74}\\
				\rowstyle{\color{blue}}
				\basePlain & \CNN & 33.12 $\pm$ 0.18  & 65.23 $\pm$ 0.18 \\
				\rowstyle{\color{blue}}
				\baseFair & \CNN & 50.21 $\pm$ 0.20 & 69.02 $\pm$ 0.18\\
				\rowstyle{\color{blue}}
				\baseBest & \CNN & 51.02 $\pm$ 0.20  &  68.98 $\pm$ 0.18\\
				\midrule
				\rowstyle{\color{blue}}
				\basePlain & \ResNet-10 & 58.60 $\pm$ 0.22& 79.99 $\pm$ 0.16\\
				\rowstyle{\color{blue}}
				\baseFair & \ResNet-10 & 64.58 $\pm$ 0.23 & 82.31 $\pm$ 0.16 \\
				\rowstyle{\color{blue}}
				\baseBest & \ResNet-10 & 65.37 $\pm$ 0.22 &  81.84 $\pm$ 0.16 \\
				\rowstyle{\color{blue}}
				\basePlain & \ResNet-18 & 62.69 $\pm$ 0.22 & 83.27 $\pm$ 0.16\\
				\rowstyle{\color{blue}}
				\baseFair & \ResNet-18 & 68.64 $\pm$ 0.22 & 84.47 $\pm$ 0.16 \\
				\rowstyle{\color{blue}}
				\baseBest & \ResNet-18 & \textbf{69.09 $\pm$ 0.22}&  \textbf{84.58 $\pm$ 0.16} \\
				\midrule
				Meta SGD~\cite{li2017meta}$^\dagger$& WRN & 62.95 $\pm$ 0.03& 79.34 $\pm$ 0.06\\
				LEO~\cite{rusu2019meta} & WRN & 66.33 $\pm$ 0.05 & 81.44 $\pm$ 0.09\\
				\rowstyle{\color{blue}}
				\basePlain & WRN & 63.85 $\pm$ 0.21 & 84.17 $\pm$ 0.15 \\
				\rowstyle{\color{blue}}
				\baseFair & WRN & 66.86 $\pm$ 0.21 & \textbf{85.50 $\pm$ 0.14} \\
				\rowstyle{\color{blue}}
				\baseBest & WRN & \textbf{69.75 $\pm$ 0.20} & 85.31 $\pm$ 0.15 \\
				\midrule
				\rowstyle{\color{blue}}
				\basePlain & MobileNet & 63.65 $\pm$ 0.22& 84.01 $\pm$ 0.16\\
				\rowstyle{\color{blue}}
				\baseFair & MobileNet & 68.66 $\pm$ 0.23& \textbf{85.43} $\pm$ 0.15\\
				\rowstyle{\color{blue}}
				\baseBest & MobileNet & \textbf{69.47 $\pm$ 0.22} & 85.17 $\pm$ 0.15\\
				\midrule
				\rowstyle{\color{blue}}
				\basePlain & DenseNet & 64.35 $\pm$ 0.23 & 85.69 $\pm$ 0.15\\
				\rowstyle{\color{blue}}
				\baseFair & DenseNet & 69.91 $\pm$ 0.22& 86.42 $\pm$ 0.15\\
				\rowstyle{\color{blue}}
				\baseBest & DenseNet & \textbf{71.32 $\pm$ 0.22}& \textbf{86.66 $\pm$ 0.15}\\
				\bottomrule
			\end{tabular}
		}
		\begin{flushleft}
			$^\dagger$: Results reported in~\cite{rusu2019meta}.\hspace{5pt}
			$^\sharp$: Results reported in~\cite{liu2019learning}.\\
			$^\star$: \cite{rusu2019meta} and our results are averaged over 10,000 rounds.\\
		\end{flushleft}
	}
\end{table}

\subsection{Results}
Table~\ref{tbMain_1}, \ref{tbMain_2}, and~\ref{tbMain_3} present our results on \emph{mini}ImageNet, \emph{tiered}ImageNet, and CIFAR-100, respectively. In line with prior work, we observe that nearest-neighbor classifiers using ``vanilla'' Euclidean distance (UN) do not perform very well. However, simply applying L2-normalization (L2N) consistently leads to accuracy gains of at least $3\%$ on these datasets. Subtracting the mean before L2-normalization (CL2N) leads to another improvement of $1\!-\!3\%$.

Our SimpleShot nearest-neighbor / nearest-centroid classifiers achieve accuracies that are comparable with or better than the state-of-the-art. For example, on the \emph{mini}ImageNet dataset, our simple methods obtain the highest one-shot and five-shot accuracies for three of five network architectures. 

\begin{table}[t]
	\centering
	\small {
		\caption{\small Average accuracy (in \%; measured over 600/10,000 rounds$^\star$) of one-shot and five-shot classifiers for five-way classification on CIFAR-100; higher is better. The best result is in \textbf{bold}  font. Results of our approach are in {\color{blue} blue}. Best viewed in color.} \label{tbMain_3}
		\resizebox{\linewidth}{!}{  
			\begin{tabular}{=l +l +l +l}
				\toprule
				\textbf{Approach} & \textbf{Network} & \textbf{One shot} & \textbf{Five shots} \\ \midrule
				TADAM~\cite{oreshkin2018tadam} & \ResNet & 40.10 $\pm$ 0.40& \textbf{56.10 $\pm$ 0.40}\\
				\rowstyle{\color{blue}}
				\basePlain & \ResNet-10 & 36.38 $\pm$ 0.17 & 52.67 $\pm$ 0.18 \\
				\rowstyle{\color{blue}}
				\baseFair & \ResNet-10 & 38.47 $\pm$ 0.17  & 53.34 $\pm$ 0.18 \\
				\rowstyle{\color{blue}}
				\baseBest & \ResNet-10 & \textbf{40.13 $\pm$ 0.18}  & 53.63 $\pm$ 0.18 \\
				\bottomrule
			\end{tabular}
		}
		\begin{flushleft}
			$^\star$: Our results are averaged over 10,000 rounds.\\
		\end{flushleft}
	}
\end{table} 


We perform a simple experiment measuring the effectiveness of feature transformations at various stages of convolutional-network training. We train a DenseNet on \emph{mini}ImageNet for 90 epochs, and measure the one-shot five-way accuracy on 10,000 tasks sampled from the validation classes after each epoch. The results of this experiment are shown in Figure~\ref{fig:concept}: they show that nearest-neighbor classifiers using C2LN feature transformation consistently outperform their UN and L2N counterparts. This suggests that our observations on the role of feature transformations do not depend on how long the network is trained.

We also investigate the effect of feature transformations on more complex few-shot learning algorithms. Specifically, we trained a \CNN architecture with the ProtoNet~\cite{snell2017prototypical} loss, which uses unnormalized Euclidean distances. After training, we apply feature transformations before computing pairwise Euclidean distances between features in a nearest-neighbor approach. Table~\ref{tb_proto} presents the results of this experiment, which shows that CL2N normalization can also improve the performance of ProtoNet.

\begin{table}
	\centering
	\small {
		\caption{\small \textbf{Feature transformations matter in 1NN classification with ProtoNet~\cite{snell2017prototypical}.}
		We report average accuracy (in \%; measured over 10,000 rounds) of five-way one-shot / five-shot ProtoNet classifiers on \emph{mini}ImageNet with and without feature transformations (applied after training).} \label{tb_proto}
		\vskip -7.5pt
		\resizebox{\linewidth}{!}{
			\begin{tabular}{+l +l +l +l}
				\toprule
				    {\color{black} \textbf{1NN (UN)~\cite{snell2017prototypical}}} & \textbf{1NN (UN; ours)} & \textbf{1NN (L2N)} & \textbf{1NN (CL2N)} \\
				\midrule
				 {\color{black}49.42 / 68.20} & 49.56 / 67.79 & 49.55 / 67.84 & \textbf{50.12 / 68.51}\\
				\bottomrule
			\end{tabular}
		}
	}
\end{table}

%% file: discussion.tex
\section{Conclusion}
\label{sec:discussion}

We analyzed the effect of simple feature transformations in nearest-neighbor classifiers for few-shot learning. We observed that such transformations --- in particular, a combination of centering and L2-normalization --- can improve the quality of the representation to a degree that the resulting classifiers outperforms several state-of-the-art approaches to few-shot learning. We hope that the SimpleShot classifiers studied in this paper will be used as a competitive baseline in future studies on few-shot learning.

%% file: suppl_content.tex
\section{Meta-iNat Results}
We also investigate the role of feature transformations in SimpleShot on the long-tailed  iNaturalist dataset~\cite{van2018inaturalist}. Following the meta-iNat benchmark~\cite{wertheimer2019few}, we split the dataset to have 908 base classes and 227 novel classes. We follow the evaluation setup of~\cite{wertheimer2019few} and perform 227-way multi-shot evaluation. (In the meta-iNat benchmark, the number of shots varies per class.)  We train all networks for 90 epochs using stochastic gradient descent. We set the initial learning rate to be $0.1$ and batch size to be $256$. We scale the learning rate by $0.1$ after every $30$ epochs.

\begin{figure}[htb!]
	\centering
	\begin{subfigure}[b]{\linewidth}
	\centering
		\includegraphics[width=\linewidth]{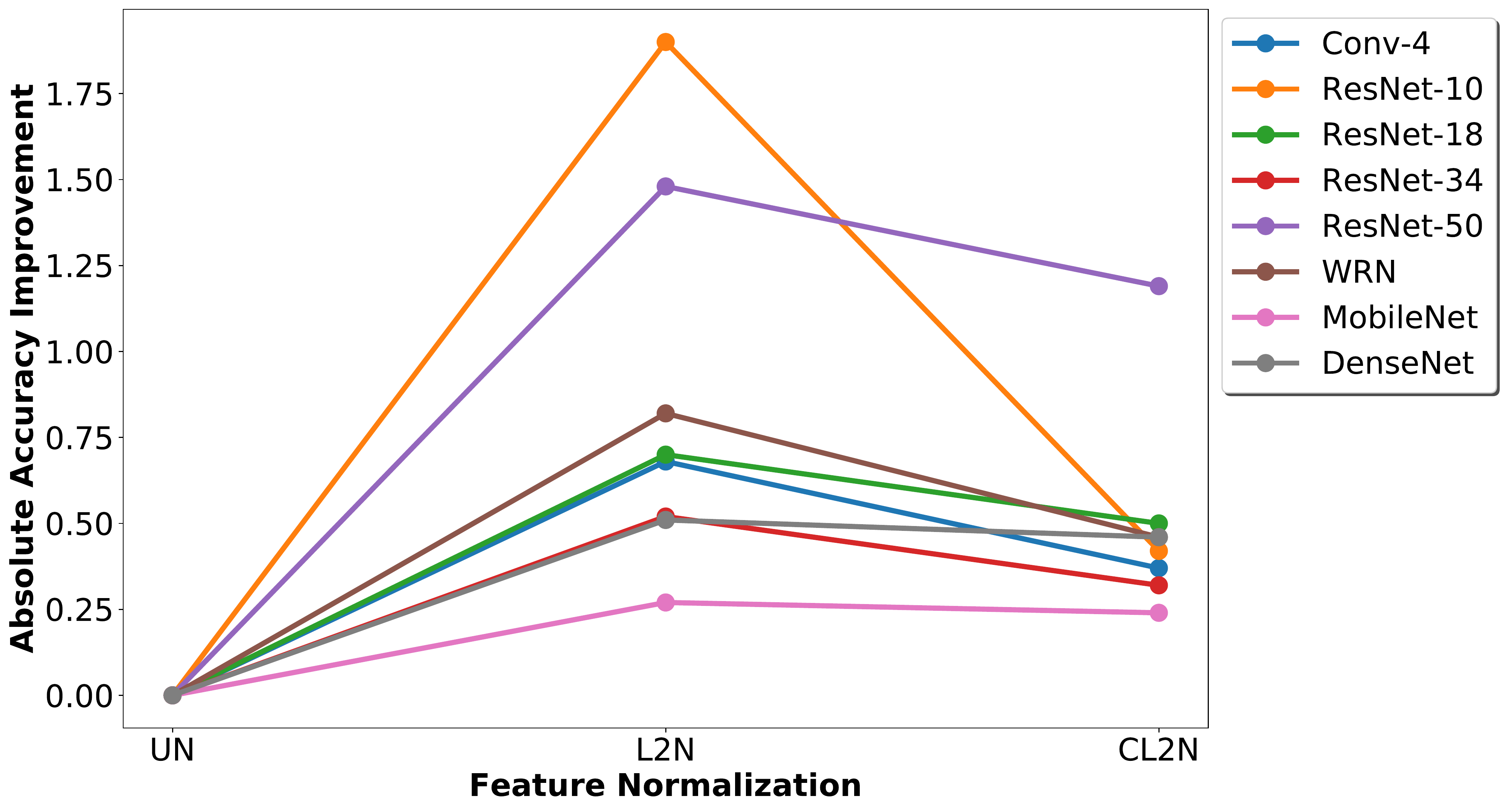}
	\end{subfigure}
	\caption{Absolute accuracy improvement (per class; in \%) on the meta-iNat dataset of SimpleShot classifiers with L2-normalization (L2N) and centering and L2-normalization (CL2N) compared to a SimpleShot classifier without feature normalization (UN).}
	\label{fig:inat}
\end{figure}

The results of our meta-iNat experiments with SimpleShot are presented in Table~\ref{table:inatural}. The table reports the averaging the accuracy on each class over all test classes (per class) and the average accuracy over all test images (mean). To the best of our knowledge, our highest accuracy of 62.13\% (per class) and 65.09\% (mean) is the current state-of-the-art on the meta-iNat benchmark. Figure~\ref{fig:inat} shows the absolute accuracy improvement (in \%) of each of the classifiers compared to the baseline nearest-neighbor classifier without feature normalization (UN). In line with prior experiments, L2-normalization (L2N) leads to accuracy improvements in few-shot learning. Different from the other experiments, centering after L2-normalization (CL2N) does not improve the accuracy of SimpleShot further.

\vfill 
\break

\begin{table}[H]
\caption{Accuracy (in \%) of SimpleShot classifiers in 227-way multi-shot classification on the meta-iNat benchmark~\cite{wertheimer2019few}. Accuracy is measured by averaging the accuracy on each class over all test classes (per class) and by averaging accuracy over all test images (mean). Higher is better.}
\label{table:inatural}
\centering
\scalebox{0.8}{
\begin{tabular}{lcccccc}

\toprule
            &    \multicolumn{2}{c}{\bf SimpleShot (UN)} &  \multicolumn{2}{c}{\bf SimpleShot (L2N)} &  \multicolumn{2}{c}{\bf SimpleShot (CL2N)}\\

            &  \bf Per class &   \bf Mean &  \bf Per class &   \bf Mean &  \bf Per class &  \bf Mean \\
\midrule
      \bf Conv-4 &  21.32 &  22.93 &        22.00 &  23.73 &       21.69 &  23.21  \\
   \bf ResNet-10 &  40.50 &  42.06 &        42.40 &  43.86 &       40.92 &  42.19  \\
   \bf ResNet-18 &  55.33 &  58.06 &        56.03 &  58.50 &       55.83 &  58.33  \\
   \bf ResNet-34 &  59.98 &  62.43 &        60.50 &  62.65 &       60.30 &  62.50  \\
   \bf ResNet-50 &  54.13 &  56.85 &        55.61 &  57.77 &       55.32 &  57.47  \\
   \bf WRN       &  60.48 &  63.22 &        61.30 &  63.77 &       60.94 &  63.42  \\
\bf MobileNet &     52.01 &  53.92 &        52.28 &  54.06 &       52.25 &  54.01  \\
\bf DenseNet &      61.62 &  64.77 &        62.13 &  65.09 &       62.08 &  65.02  \\
 \bottomrule

\end{tabular}}

\end{table}


%% file: main.bbl
\begin{thebibliography}{10}\itemsep=-1pt

\bibitem{bauer2017discriminative}
M.~Bauer, M.~Rojas-Carulla, J.~B. Swiatkowski, B.~Scholkopf, and R.~E. Turner.
\newblock Discriminative k-shot learning using probabilistic models.
\newblock {\em arXiv preprint arXiv:1706.00326}, 2017.

\bibitem{chen2019closer}
W.-Y. Chen, Y.-C. Liu, Z.~Kira, Y.-C.~F. Wang, and J.-B. Huang.
\newblock A closer look at few-shot classification.
\newblock In {\em ICLR}, 2019.

\bibitem{fei2006one}
L.~Fei-Fei, R.~Fergus, and P.~Perona.
\newblock One-shot learning of object categories.
\newblock {\em PAMI}, 28(4):594--611, 2006.

\bibitem{finn2017model}
C.~Finn, P.~Abbeel, and S.~Levine.
\newblock Model-agnostic meta-learning for fast adaptation of deep networks.
\newblock In {\em ICML}, 2017.

\bibitem{finn2018probabilistic}
C.~Finn, K.~Xu, and S.~Levine.
\newblock Probabilistic model-agnostic meta-learning.
\newblock In {\em NeurIPS}, 2018.

\bibitem{garcia2018few}
V.~Garcia and J.~Bruna.
\newblock Few-shot learning with graph neural networks.
\newblock In {\em ICLR}, 2018.

\bibitem{gauthier1998}
I.~Gauthier.
\newblock {\em Dissecting face recognition: The role of expertise and level of
  categorization in object recognition}.
\newblock PhD thesis, Yale University, 1998.

\bibitem{gidaris2018dynamic}
S.~Gidaris and N.~Komodakis.
\newblock Dynamic few-shot visual learning without forgetting.
\newblock In {\em CVPR}, 2018.

\bibitem{gordon2019meta}
J.~Gordon, J.~Bronskill, M.~Bauer, S.~Nowozin, and R.~Turner.
\newblock Meta-learning probabilistic inference for prediction.
\newblock In {\em ICLR}, 2019.

\bibitem{grant2018recasting}
E.~Grant, C.~Finn, S.~Levine, T.~Darrell, and T.~Griffiths.
\newblock Recasting gradient-based meta-learning as hierarchical bayes.
\newblock In {\em ICLR}, 2018.

\bibitem{he2016deep}
K.~He, X.~Zhang, S.~Ren, and J.~Sun.
\newblock Deep residual learning for image recognition.
\newblock In {\em CVPR}, 2016.

\bibitem{vanhorn2017}
G.~V. Horn and P.~Perona.
\newblock The devil is in the tails: Fine-grained classification in the wild.
\newblock In {\em arXiv 1709.01450}, 2017.

\bibitem{howard2017mobilenets}
A.~G. Howard, M.~Zhu, B.~Chen, D.~Kalenichenko, W.~Wang, T.~Weyand,
  M.~Andreetto, and H.~Adam.
\newblock Mobilenets: Efficient convolutional neural networks for mobile vision
  applications.
\newblock {\em arXiv preprint arXiv:1704.04861}, 2017.

\bibitem{huang2017densely}
G.~Huang, Z.~Liu, L.~Van Der~Maaten, and K.~Q. Weinberger.
\newblock Densely connected convolutional networks.
\newblock In {\em CVPR}, 2017.

\bibitem{jiang2019learning}
X.~Jiang, M.~Havaei, F.~Varno, G.~Chartrand, N.~Chapados, and S.~Matwin.
\newblock Learning to learn with conditional class dependencies.
\newblock In {\em ICLR}, 2019.

\bibitem{krizhevsky2009learning}
A.~Krizhevsky and G.~Hinton.
\newblock Learning multiple layers of features from tiny images.
\newblock Technical report, University of Toronto, 2009.

\bibitem{lee2018meta}
Y.~Lee and S.~Choi.
\newblock Gradient-based meta-learning with learned layerwise metric and
  subspace.
\newblock In {\em ICML}, 2018.

\bibitem{li2017meta}
Z.~Li, F.~Zhou, F.~Chen, and H.~Li.
\newblock Meta-sgd: Learning to learn quickly for few shot learning.
\newblock {\em arXiv preprint arXiv:1707.09835}, 2017.

\bibitem{liu2019learning}
Y.~Liu, J.~Lee, M.~Park, S.~Kim, E.~Yang, S.~J. Hwang, and Y.~Yang.
\newblock Learning to propagate labels: Transductive propagation network for
  few-shot learning.
\newblock In {\em Proceedings of the Annual Meeting of the Cognitive Science
  Society}, 2019.

\bibitem{Miller95}
G.~A. Miller.
\newblock Wordnet: a lexical database for english.
\newblock {\em Communications of the ACM}, 38(11):39--41, 1995.

\bibitem{mishra2018simple}
N.~Mishra, M.~Rohaninejad, X.~Chen, and P.~Abbeel.
\newblock A simple neural attentive meta-learner.
\newblock In {\em ICLR}, 2018.

\bibitem{munkhdalai2018rapid}
T.~Munkhdalai, X.~Yuan, S.~Mehri, and A.~Trischler.
\newblock Rapid adaptation with conditionally shifted neurons.
\newblock In {\em ICML}, 2018.

\bibitem{nichol2018first}
A.~Nichol, J.~Achiam, and J.~Schulman.
\newblock On first-order meta-learning algorithms.
\newblock {\em CoRR, abs/1803.02999}, 2018.

\bibitem{oreshkin2018tadam}
B.~N. Oreshkin, A.~Lacoste, and P.~Rodriguez.
\newblock Tadam: Task dependent adaptive metric for improved few-shot learning.
\newblock In {\em NeurIPS}, 2018.

\bibitem{qiao2018few}
S.~Qiao, C.~Liu, W.~Shen, and A.~L. Yuille.
\newblock Few-shot image recognition by predicting parameters from activations.
\newblock In {\em CVPR}, 2018.

\bibitem{ravi2017optimization}
S.~Ravi and H.~Larochelle.
\newblock Optimization as a model for few-shot learning.
\newblock In {\em ICLR}, 2017.

\bibitem{ren2018meta}
M.~Ren, E.~Triantafillou, S.~Ravi, J.~Snell, K.~Swersky, J.~B. Tenenbaum,
  H.~Larochelle, and R.~S. Zemel.
\newblock Meta-learning for semi-supervised few-shot classification.
\newblock In {\em ICLR}, 2018.

\bibitem{RussakovskyDSKS15ImageNet}
O.~Russakovsky, J.~Deng, H.~Su, J.~Krause, S.~Satheesh, S.~Ma, Z.~Huang,
  A.~Karpathy, A.~Khosla, M.~S. Bernstein, A.~C. Berg, and F.-F. Li.
\newblock Imagenet large scale visual recognition challenge.
\newblock {\em IJCV}, 115(3):211--252, 2015.

\bibitem{rusu2019meta}
A.~A. Rusu, D.~Rao, J.~Sygnowski, O.~Vinyals, R.~Pascanu, S.~Osindero, and
  R.~Hadsell.
\newblock Meta-learning with latent embedding optimization.
\newblock In {\em ICLR}, 2019.

\bibitem{snell2017prototypical}
J.~Snell, K.~Swersky, and R.~Zemel.
\newblock Prototypical networks for few-shot learning.
\newblock In {\em NeurIPS}, 2017.

\bibitem{sung2018learning}
F.~Sung, Y.~Yang, L.~Zhang, T.~Xiang, P.~H. Torr, and T.~M. Hospedales.
\newblock Learning to compare: Relation network for few-shot learning.
\newblock In {\em CVPR}, 2018.

\bibitem{triantafillou2017few}
E.~Triantafillou, R.~Zemel, and R.~Urtasun.
\newblock Few-shot learning through an information retrieval lens.
\newblock In {\em CVPR}, 2017.

\bibitem{van2018inaturalist}
G.~Van~Horn, O.~Mac~Aodha, Y.~Song, Y.~Cui, C.~Sun, A.~Shepard, H.~Adam,
  P.~Perona, and S.~Belongie.
\newblock The inaturalist species classification and detection dataset.
\newblock In {\em Proceedings of the IEEE conference on computer vision and
  pattern recognition}, pages 8769--8778, 2018.

\bibitem{vinyals2016matching}
O.~Vinyals, C.~Blundell, T.~Lillicrap, D.~Wierstra, et~al.
\newblock Matching networks for one shot learning.
\newblock In {\em NIPS}, 2016.

\bibitem{wertheimer2019few}
D.~Wertheimer and B.~Hariharan.
\newblock Few-shot learning with localization in realistic settings.
\newblock In {\em Proceedings of the IEEE Conference on Computer Vision and
  Pattern Recognition}, pages 6558--6567, 2019.

\bibitem{ye2018learning}
H.-J. Ye, H.~Hu, D.-C. Zhan, and F.~Sha.
\newblock Learning embedding adaptation for few-shot learning.
\newblock {\em arXiv preprint arXiv:1812.03664}, 2018.

\bibitem{yoon2018bayesian}
J.~Yoon, T.~Kim, O.~Dia, S.~Kim, Y.~Bengio, and S.~Ahn.
\newblock Bayesian model-agnostic meta-learning.
\newblock In {\em NeurIPS}, 2018.

\bibitem{zagoruyko2016wide}
S.~Zagoruyko and N.~Komodakis.
\newblock Wide residual networks.
\newblock In {\em BMVC}, 2016.

\end{thebibliography}
